\documentclass{article}
\usepackage{iclr2026_conference,times}

\usepackage{amsmath,amsfonts,bm}

\def\eqref#1{equation~\ref{#1}}

\def\1{\bm{1}}

\DeclareMathAlphabet{\mathsfit}{\encodingdefault}{\sfdefault}{m}{sl}
\SetMathAlphabet{\mathsfit}{bold}{\encodingdefault}{\sfdefault}{bx}{n}

\usepackage{hyperref}
\usepackage{url}

\usepackage{graphicx}
\usepackage{booktabs}
\usepackage{multirow}

\title{Zero-shot Multivariate Time Series Forecasting Using Tabular Prior Fitted Networks}

\author{Mayuka Jayawardhana$^{1}$, Nihal Sharma$^{2}$, Kazem Meidani$^{2}$, Bayan Bruss$^{2}$, Tom Goldstein$^{1}$ \\
\textbf{\& Doron Bergman}$^{2}$ \thanks{Corresponding author} \\
$^{1}$University of Maryland, $^{2}$Capital One\\
\texttt{\{mayukaj,tomg\}@umd.edu} \\
\texttt{\{nihal.sharma,mohammadkazem.meidani\}@capitalone.com} \\
\texttt{\{bayan.bruss,doron.bergman\}@capitalone.com}
}

\iclrfinalcopy

\begin{document}

\maketitle

\begin{abstract}

Tabular foundation models, particularly Prior-data Fitted Networks like TabPFN have emerged as the leading contender in a myriad of tasks ranging from data imputation to label prediction on the tabular data format surpassing the historical successes of tree-based models. This has led to investigations on their applicability to forecasting time series data which can be formulated as a tabular problem. While recent work to this end has displayed positive results, most works have limited their treatment of multivariate time series problems to several independent univariate time series forecasting subproblems, thus ignoring any inter-channel interactions. Overcoming this limitation, we introduce a generally applicable framework for multivariate time series forecasting using tabular foundation models. We achieve this by recasting the multivariate time series forecasting problem as a \emph{series} of scalar regression problems which can then be solved \emph{zero-shot} by any tabular foundation model with regression capabilities. We present results of our method using the TabPFN-TS backbone and compare performance with the current state of the art tabular methods.

\end{abstract}

\section{Introduction}

Multivariate time series (MTS) are ubiquitous in real-world applications, presenting themselves as critical regression, classification, or forecasting tasks across domains such as finance, meteorology, and engineering \cite{10.1098/rsta.2020.0209, Ismail_Fawaz_2019, GVK483463442, duchon2012time}. Unlike their univariate counterparts, MTS data exhibit added complexity due to the inclusion of both temporal (inter-sample) and spatial (intra-sample) dependencies. While it is reasonable to assume that best possible forecasting necessitates modeling the joint probability distribution of these evolving variables, prior work has often sidestepped the complexity of intra-sample dependence among covariates. Instead, many approaches simply decompose MTS into independent univariate problems, a strategy often referred to as Channel Independence (CI) \cite{han2024capacity, wu2021autoformer}. Reformulating the problem is particularly appealing in regimes where training data is not abundant --- modeling both Channel Dependent (intra-sample) and CI interactions generally would need models with higher capacity and thus, more data \cite{han2024capacity}. 

This CI decomposition of MTS problems is also appealing since it makes MTS tasks organically applicable to be solved using a class models that are trained to regress over a scalar target variable using data provided in the context (a.k.a In-Context Learning). Particularly, Prior-data Fitted Networks (PFNs) are trained to be in-context learners for general tabular data: given all information in a table apart from the label of one of the rows, these networks learn to estimate this scalar target label. While the training data here assumes that every row of a table is independently generated given a fixed underlying dependency structure, prior works have suggested the use of PFNs for univariate time series forecasting by treating the future value of the time series as the target label and have been found to be competitive with purely temporal models. This observation has led to two streams of work for MTS forecasting: one that involves data and architectural modifications to adapt the PFN-like training to account for the temporal dependence (\cite{dooley2023forecastpfn, moroshan2025tempopfn} for univariate forecasting, \cite{taga2025timepfn,ansari2025chronos2} for MTS problems), and a second, that includes zero-shot methods which reformulate the univariate time series forecasting problem to appear as a tabular regression task by augmenting the time series data to fit the input data format of tabular foundation models \cite{hoo2025tablestimetabpfnv2outperforms}.

\begin{figure}[t]
    \centering
    \includegraphics[width=\linewidth]{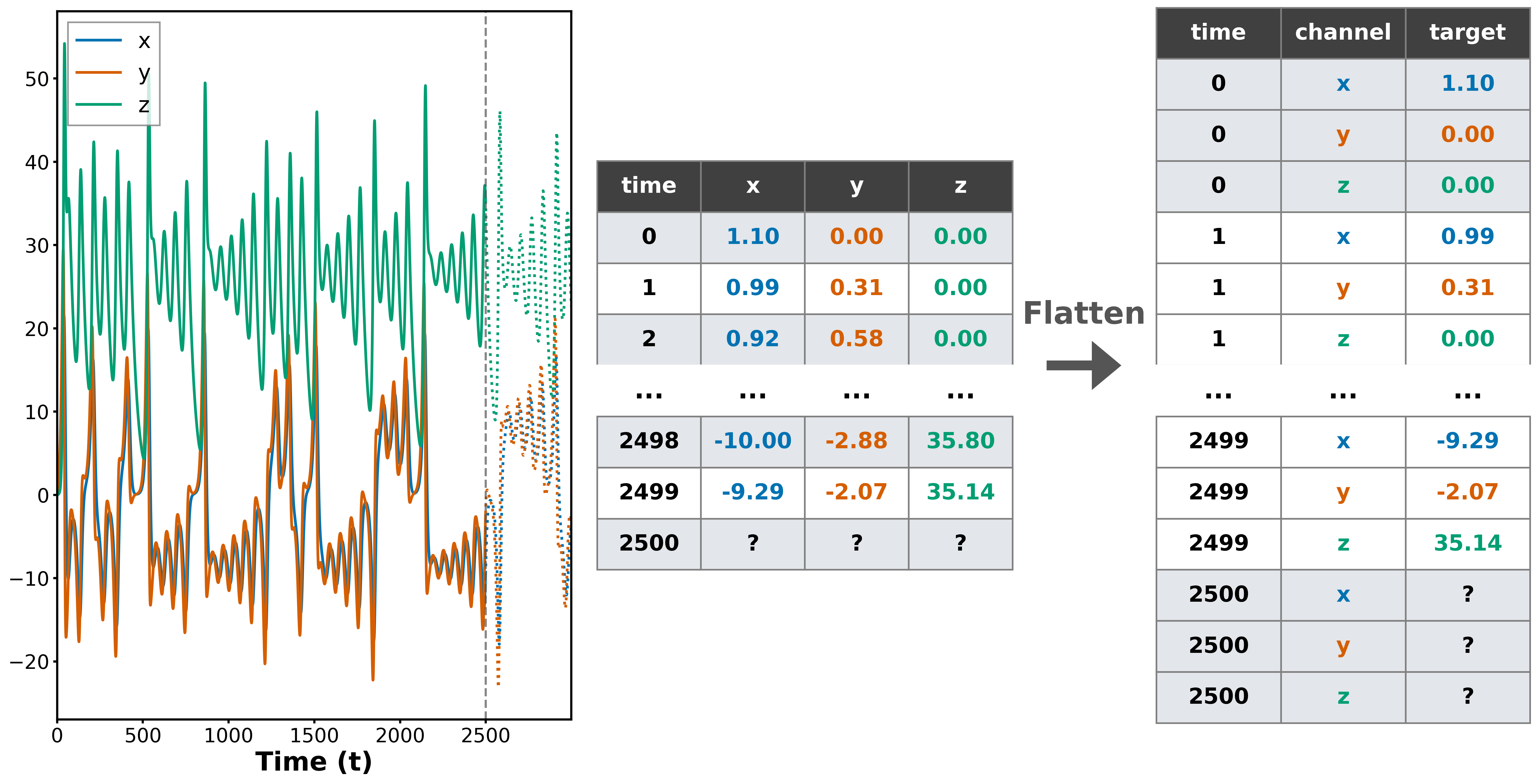}
    \caption{\textbf{Reformulating multivariate time series forecasting as a scalar regression problem}: A pictorial and tabular representation of a Lorenz system (a time series with three covariates $x,y,z$ introduced in \cite{DeterministicNonperiodicFlow}) is presented on the left parts of the figure. The time-indexed table is ``rolled out'' by introducing a channel indicator column (as seen in the table on the right). We note that this transformation expands the size of the table by a factor equal to the number of covariates.}
    \label{fig:lorenz}
    \vspace{-0.2cm}
\end{figure}

In this work, we propose a new framework that reformulates MTS forecasting tasks to fit the mold of the input data of tabular foundation models and can be used to extract zero-shot predictions of future samples. Building on the existing technique of univariate time series data augmentation, our method explicitly models the intra-sample dependencies among covariates without requiring architectural modifications or retraining. We achieve this by serializing the multivariate structure into a ``rolled out'' tabular format: at each time step, the multivariate vector is flattened into several rows, where the features consist of the timestamp, the covariate index, and the value (see Figure \ref{fig:lorenz}). The regression target is structured to be the subsequent value in this flattened sequence. This method effectively transforms the spatial correlations between variables into a sequential dependency that the PFN can interpret. Our method trades off the zero-shot applicability of off-the-shelf tabular foundation models against increased computational costs, which arise from the inflated context lengths inherent in unrolling the MTS data.

We empirically evaluate our approach by benchmarking it against the univariate decomposition baseline established by TabPFN-TS \cite{hoo2025tablestimetabpfnv2outperforms}, as well as other specialized architectures trained on time series priors, including TimePFN \cite{taga2025timepfn}, TempoPFN \cite{moroshan2025tempopfn}, and Chronos-2 \cite{ansari2025chronos2}. The remainder of this paper is organized as follows: Section 2 details our methodology; Section 3 presents our experimental results and comparative analysis; and Section 4 provides the conclusion and directions for future work.

\section{Methodology}
In this section we discuss our lightweight approach for multivariate time series forecasting using tabular PFN regressors for scalar targets. TabPFN-TS \cite{hoo2025tablestimetabpfnv2outperforms} demonstrates that standard tabular PFNs can solve time series forecasting problems by treating them as regression problems - time is the independent variable, and the time series value is the dependent variable. Additionally, augmenting the time series data with simple auto generated temporal features, as additional independent variables, is sufficient to make the performance of TabPFN-TS in time series forecasting competitive. However, TabPFN-TS does not have a mechanism for unconstrained multivariate forecasting that models cross-channel dependencies. We demonstrate that a simple reformulation strategy enables efficient multivariate time series forecasting that can utilize cross-channel context using tabular PFNs.

\subsection{Univariate time series forecasting as a regression problem}

Time series forecasting can be stated as predicting the distribution of a horizon $H$ of next few observations, $x_{t:t+H} = \{x_t, x_{t+2}, \dots, x_{t+H - 1}\}$, given all past observations $x_{<t} = x_{0:t} = \{x_0, x_1, \dots, x_{t-1}\}$: $P(x_{t:t+H} | x_{<t})$. The joint probability of a time series sequence $\{x_0, \dots, x_T\}$ can be decomposed into a product of conditional next-step predictions: $P(x_0, \dots, x_T) = \prod_{t=0}^T P(x_t | x_{<t})$. Therefore, the fundamental task is solving the next-step prediction $P(x_t | x_{<t})$ for all $t$. We can frame $P(x_t | x_{<t})$ as a standard conditional density estimation (regression) problem: $P(x_t| x_{<t}) =  P(u| t; (t_i,u_i)|_{i=1}^{t-1})$ .

By defining the feature vector to include time as a explicit feature, a regression model can leverage $t$ as the sample feature to predict target $y=x_t$. Tabular PFNs are trained to solve regression problems in this form.

\subsection{Multivariate time series forecasting as a univariate regression problem}

Consider the simplest multivariate time series: A two variate time series forecasting problem can be stated as $P(x_{t}, y_{t} | x_{<t}, y_{<t})$.

Existing tabular PFN regressors are trained to predict only one target variable. As re-training a foundational model that can natively predict multiple targets requires considerable resources and entails engineering challenges, we employ the following simple reformulation of the problem. 

We can introduce an indicator variable $\eta \in \{x, y\}$ to denote which component is being predicted. Features for each sample are now: $t, \eta$ and we predict the single target variable $u$. The prediction problem can now be reformulated as,

\begin{subequations}
\begin{align}
    P(x_t|x_{<t}, y_{<t}) = & P(u|\eta=x,t;(\eta = x, s; u_s) |_{s=1}^{t-1}, (\eta = y, s; u_s) |_{s=1}^{t-1}) \\
    P(y_t|x_t, x_{<t}, y_{<t}) = & P(u|\eta=y,t;(\eta = x, s; u_s) |_{s=1}^{t}, (\eta = y, s; u_s) |_{s=1}^{t-1})
\end{align}
\end{subequations}

Predicting the different components autoregressivly draws predictions from the joint multivariate distribution of next steps in the multivariate timeseries, but the joint distribution $P(x_{t}, y_{t} | x_{<t}, y_{<t})$ can also be drawn from directly, as we do in our implementation.

More generally we can simply use the following notation to describe the multivariate timeseries,
$
   P(u_{a,t}|u_{b,<t}|_{b=1}^d) = P(u_|a,t;(b, t_i, u_{b,i})|_{i=1, b=1}^{i=t-1, b=d}) 
$

While our method introduces a position-dependent indicator column ($\eta$), we do not experiment with shuffling its order (as detailed in Appendix \ref{sec:sensitivity}). This formulation is easily extended to more variates by treating $\eta$ as a categorical feature which encodes which variate that row treats as the prediction target, and both the different channels and the full forecasting horizon can all be predicted simultaneously using any tabular foundation model.  

In practice, before combining the channels to form a unified sequence we normalize each channel's sequence independently (z-score normalization) before passing it to TabPFN. The predictions are then mapped back to the original channel scale, by applying the inverse to the z-score normalization before calculating metrics. We find that this independent normalization helps when different variates are of different scales, which is a common occurrence (see Appendix \ref{sec:mitigation} for more details). 

Thus, this simple feature engineering approach enables the use of off-the shelf tabular foundational models for multivariate forecasting. In what follows we label our approach as \textbf{TabPFN-TS-MV}.
\section{Experiments}

We evaluate our approach using the Gift-Eval time series benchmark \cite{aksu2024giftevalbenchmarkgeneraltime}. Gift-Eval spans 23 datasets with diverse characteristics, totaling over 144,000 time series and 177 million data points across seven application domains and ten sampling frequencies. It includes both univariate and multivariate forecasting settings and covers prediction horizons ranging from short- to long-term. In total, combining all valid datasets, frequencies, and horizons, GIFT-Eval provides 97 distinct benchmarking tasks. As our method differs from TabPFN-TS on multivariate time series we only compare the multivariate datasets. We note that our method is in principle sensitive to the order and choice of variates used for forecasting. We do not perform ablations on these as the order of variates is inherently permuted by TabPFN-TS during inference and the effects of censoring the participation of some variates would be dataset-specific and not fundamental to the applicability of our method.

\subsection{Point Forecast Accuracy}
\begin{figure}[h]
    \centering
    \includegraphics[width=\linewidth]{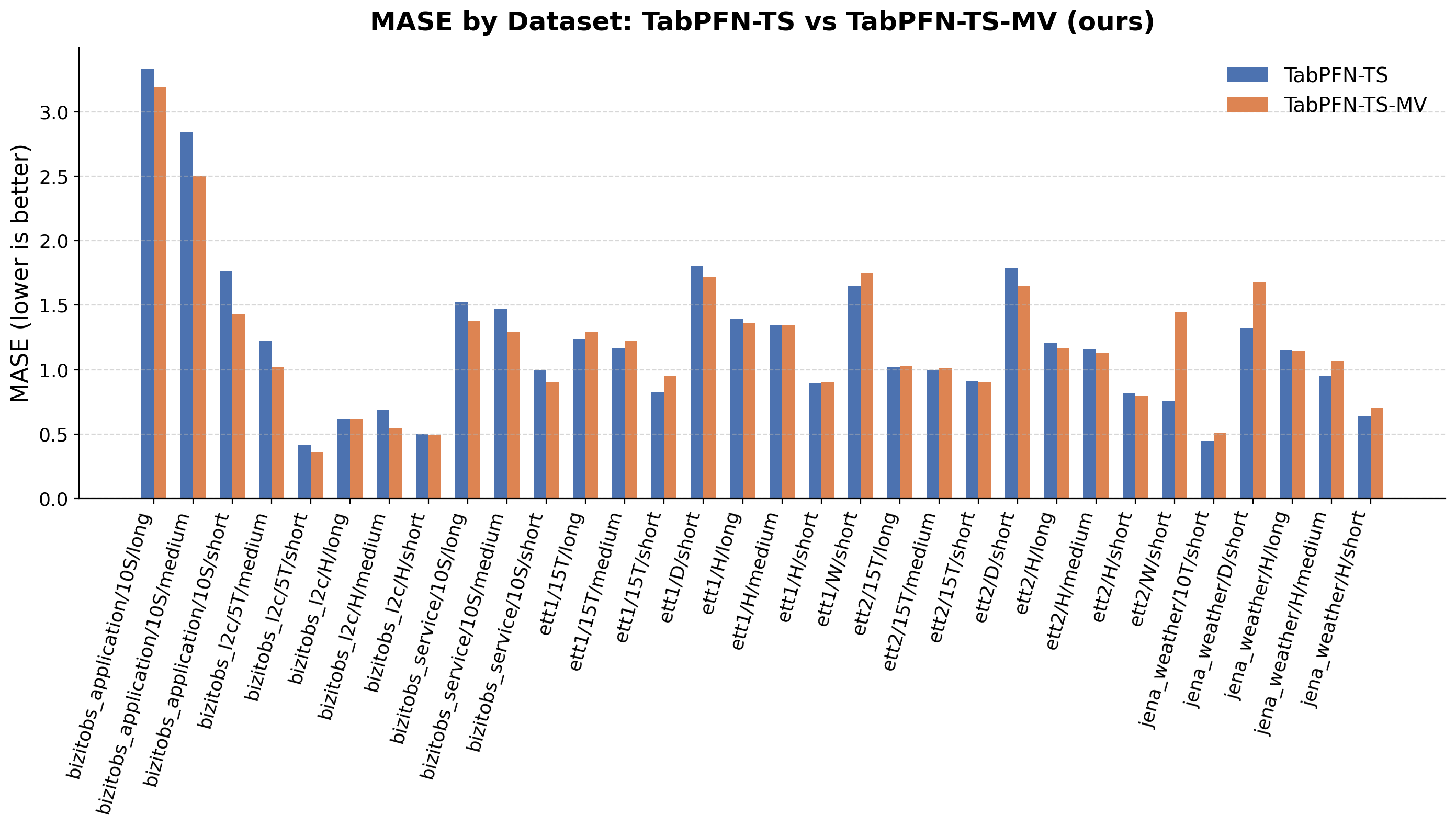}
    \caption{\textbf{Performance of TabPFN-TS compared to ours on multivariate datasets in the gift-eval benchmark} our approach lowers MASE on 60\% of the datasets. We refer the reader to Table \ref{tab:full_results} in the appendix for detailed results.}
    \label{fig:mase_indicator}
\end{figure}

Following standard practice we evaluate point forecast accuracy using the Mean Absolute Scaled Error (MASE) and observe that we outperform the standard TabPFN-TS model on 60\% of multivariate datasets in Gift-Eval \ref{fig:mase_indicator}. Note that these results were generated using the same number of past time steps for both approaches. In practice the context length limits of TabPFN V2 restricts our approach and independently predicting each variate allows for the use of more time steps. However, newer tabular foundation models with expanded context capabilities, such as TabPFN 2.5 \cite{TabPFN-2.5}, will alleviate this limitation. Further comparison of our approach against SOTA time series models is provided in Appendix \ref{sec:sota}.

\section{Discussion and Future Work}

In this work, we introduced a framework for multivariate time series forecasting that can be applied in a zero-shot manner to PFN-based regressors by reformulating the task into a series of scalar regression problems. In contrast to the existing channel-independent decomposition approach, our method allows for inter-channel interaction that could aid in improving performance of forecasting at the cost of increased context length. While this increased cost is not significant for the datasets in the Gift-Eval benchmark, we note that this might be an important consideration as the number of variates in the time-series of interest grows; the context length scales linearly in the number of variates and is also limited by maximum context length that can be used for the base tabular foundation model.

Our approach is similar in spirit to MORPHEUS \cite{patil2025morpheus}, which flattens multivariate time series into a tokenized sequence of discretized inputs for a transformer model. On the other hand, our framework is model-agnostic and requires no retraining or architectural modifications. 

While we find the empirical performance of our method to be mostly positive, we observe the channel-independent to work better under some datasets. Determining precisely when a CI or CD method is superior remains an open question in the field. We hypothesize that probabilistic forecasting using quantile regression within our framework may require better calibration, and exploring further mitigation strategies to address this deficit serves as an important direction for future research.

\bibliography{iclr2026_conference}
\bibliographystyle{iclr2026_conference}

\clearpage
\appendix
\section{Probabilistic Forecast Accuracy}\label{appendix:full_results}
\label{sec:prob_forecast}

We further evaluate probabilistic forecast accuracy via the Weighted Quantile Loss (WQL). While our standardization strategies prove beneficial (Appendix \ref{sec:mitigation}), our approach demonstrates a performance deficit compared to TabPFN-TS (Table \ref{tab:full_results}). Future work is required to isolate the drivers of this behavior and investigate potential mitigation strategies.

\section{Sensitivity of TabPFN-TS-MV to column permutations.}
\label{sec:sensitivity}
Our approach leverages the TabPFN-TS framework that introduces temporal feature engineering to TabPFN v2 \cite{hoo2025tablestimetabpfnv2outperforms}. In our experiments, we maintain a fixed column order for input data including the position of the channel indicator column. Since our method enforces an ordering across columns when the time series is rolled out, it is natural that the performance of our method is sensitive to column permutations. However, TabPFN v2 inherently ensures robustness to this ordering choice: it's inference process is an ensemble method that automatically performs row and column permutations and subsampling. Thus, we do not perform any ablations on the column ordering in this work.

\section{Detailed Results}

\begin{table}[ht]
\resizebox{!}{0.42\columnwidth}{
\begin{tabular}{lcccc}
\toprule
\multicolumn{1}{c}{\multirow{2}{*}{Dataset}} & \multicolumn{2}{c}{MASE}              & \multicolumn{2}{c}{WQL}               \\
\multicolumn{1}{c}{}                         & TabPFN-TS       & TabPFN-TS-MV (ours) & TabPFN-TS       & TabPFN-TS-MV (ours) \\
\midrule
bizitobs\_application/10S/long               & 3.3314          & \textbf{3.1884}     & \textbf{0.0483} & 0.0490              \\
bizitobs\_application/10S/medium             & 2.8469          & \textbf{2.5012}     & \textbf{0.0395} & 0.0415              \\
bizitobs\_application/10S/short              & 1.7608          & \textbf{1.4309}     & 0.0178          & \textbf{0.0171}     \\
bizitobs\_l2c/5T/medium                      & 1.2231          & \textbf{1.0165}     & 0.5009          & \textbf{0.4470}     \\
bizitobs\_l2c/5T/short                       & 0.4127          & \textbf{0.3548}     & 0.1046          & \textbf{0.0973}     \\
bizitobs\_l2c/H/long                         & 0.6166          & \textbf{0.6165}     & \textbf{0.2880} & 0.3027              \\
bizitobs\_l2c/H/medium                       & 0.6899          & \textbf{0.5444}     & 0.3120          & \textbf{0.2815}     \\
bizitobs\_l2c/H/short                        & 0.5022          & \textbf{0.4904}     & 0.2242          & \textbf{0.2241}     \\
bizitobs\_service/10S/long                   & 1.5209          & \textbf{1.3807}     & \textbf{0.0524} & 0.0527              \\
bizitobs\_service/10S/medium                 & 1.4672          & \textbf{1.2911}     & \textbf{0.0365} & 0.0409              \\
bizitobs\_service/10S/short                  & 0.9972          & \textbf{0.9042}     & 0.0181          & \textbf{0.0169}     \\
ett1/15T/long                                & \textbf{1.2373} & 1.2947              & \textbf{0.2883} & 0.3208              \\
ett1/15T/medium                              & \textbf{1.1681} & 1.2228              & \textbf{0.2766} & 0.3071              \\
ett1/15T/short                               & \textbf{0.8265} & 0.9547              & \textbf{0.1845} & 0.2301              \\
ett1/D/short                                 & 1.8064          & \textbf{1.7220}     & \textbf{0.2882} & 0.3112              \\
ett1/H/long                                  & 1.3978          & \textbf{1.3638}     & 0.2825          & \textbf{0.2782}     \\
ett1/H/medium                                & \textbf{1.3425} & 1.3480              & 0.2799          & \textbf{0.2700}     \\
ett1/H/short                                 & \textbf{0.8913} & 0.9027              & \textbf{0.1998} & 0.2033              \\
ett1/W/short                                 & \textbf{1.6502} & 1.7472              & \textbf{0.2824} & 0.2948              \\
ett2/15T/long                                & \textbf{1.0210} & 1.0254              & 0.1050          & \textbf{0.1022}     \\
ett2/15T/medium                              & \textbf{0.9968} & 1.0091              & 0.1033          & \textbf{0.1032}     \\
ett2/15T/short                               & 0.9074          & \textbf{0.9028}     & \textbf{0.0779} & 0.0804              \\
ett2/D/short                                 & 1.7877          & \textbf{1.6473}     & 0.1306          & \textbf{0.1243}     \\
ett2/H/long                                  & 1.2061          & \textbf{1.1680}     & 0.1247          & \textbf{0.1184}     \\
ett2/H/medium                                & 1.1583          & \textbf{1.1288}     & 0.1192          & \textbf{0.1133}     \\
ett2/H/short                                 & 0.8164          & \textbf{0.7941}     & 0.0720          & \textbf{0.0693}     \\
ett2/W/short                                 & \textbf{0.7571} & 1.4492              & \textbf{0.0982} & 0.1507              \\
jena\_weather/10T/short                      & \textbf{0.4473} & 0.5111              & \textbf{0.0690} & 0.0701              \\
jena\_weather/D/short                        & \textbf{1.3235} & 1.6777              & \textbf{0.0508} & 0.0786              \\
jena\_weather/H/long                         & 1.1469          & \textbf{1.1449}     & \textbf{0.0607} & 0.0714              \\
jena\_weather/H/medium                       & \textbf{0.9479} & 1.0637              & \textbf{0.0619} & 0.0642              \\
jena\_weather/H/short                        & \textbf{0.6395} & 0.7051              & \textbf{0.0481} & 0.0543              \\
\midrule
\multicolumn{1}{c}{\textbf{Average}}         & 1.2139          & \textbf{1.2032}     & \textbf{0.1514} & 0.1558    \\
\bottomrule
\end{tabular}}
\caption{\textbf{Detailed performance comparison with TabPFN-TS.} We evaluate our approach using the multivariate datsets in the Gift-Eval time series benchmark \cite{aksu2024giftevalbenchmarkgeneraltime}. Following standard practice we evaluate point forecast accuracy using the Mean Absolute Scaled Error (MASE) and probabilistic forecast accuracy via the Weighted Quantile Loss (WQL). Lower is better for both metrics. Standard deviations are omitted, as variance across random seeds for TabPFN was negligible.}
\label{tab:full_results}
\end{table}

\clearpage
\section{Comparison with SOTA Time Series Models}
\label{sec:sota}

We compare our approach against current state-of-the-art (SOTA) foundational time series models below. Consistent with our main results, average metrics are computed using the multivariate datasets from Gift-Eval. Notably, Chronos-2 exhibited significantly degraded performance on the Jena-Weather dataset, likely due to its high variate count (21). Consequently, we also report average metrics excluding this specific dataset to provide a more robust comparison.

\begin{table}[ht]
\resizebox{!}{0.084\columnwidth}{
\begin{tabular}{cccccc}
\toprule
\multicolumn{1}{l}{} & \multicolumn{5}{c}{MASE}                                                            \\
\multicolumn{1}{l}{} & Chronos 2           & TempoPFN    & TimePFN     & TabPFN-TS   & TabPFN-TS-MV (ours) \\
\midrule
Average              & 6.8068         & 1.2838 & 2.8429 & 1.2386      & \textbf{1.2255}     \\
Average (w/o jena)   & \textbf{1.0355} & 1.32823 & 3.00798 & 1.2718      & 1.2370              \\
Average Rank         & \textbf{1.6452} & 2.8387 & 4.871 & 2.9677 & 2.6774   \\
\bottomrule
\end{tabular}}
\caption{\textbf{MASE comparison with SOTA Models (Lower is better):} Average (w/o jena) denotes the average MASE across the multivariate datasets in the Gift-Eval Benchmark excluding the Jena Weather dataset \cite{jena_climate_dataset}. We denote this metric due the poor performance of Chronos-2 specifically for this dataset.}
\label{tab:mase_sota}
\end{table}

\begin{table}[ht]
\resizebox{!}{0.084\columnwidth}{
\begin{tabular}{cccccc}
\toprule
\multicolumn{1}{l}{} & \multicolumn{5}{c}{WQL}                                                                     \\
\multicolumn{1}{l}{} & Chronos 2             & TempoPFN     & TimePFN      & TabPFN-TS       & TabPFN-TS-MV (ours) \\
\midrule
Average              & 0.2577          & 0.1511 & 0.3354 & \textbf{0.1541} & 0.1586              \\
Average (w/o jena)   & \textbf{0.1392} & 0.1648  & 0.3659 & 0.1687          & 0.1721              \\
Average Rank         & \textbf{1.5484}  & 2.6129  & 4.871  & 2.9032     & 3.0645  \\
\bottomrule
\end{tabular}}
\caption{\textbf{WQL comparison with SOTA Models (Lower is better):} Average (w/o jena) denotes the average WQL across the multivariate datasets in the Gift-Eval Benchmark excluding the Jena Weather dataset \cite{jena_climate_dataset}. We denote this metric due the poor performance of Chronos-2 specifically for this dataset.}
\label{tab:wql_sota}
\end{table}

\begin{figure}[ht]
    \centering
    \includegraphics[width=\linewidth]{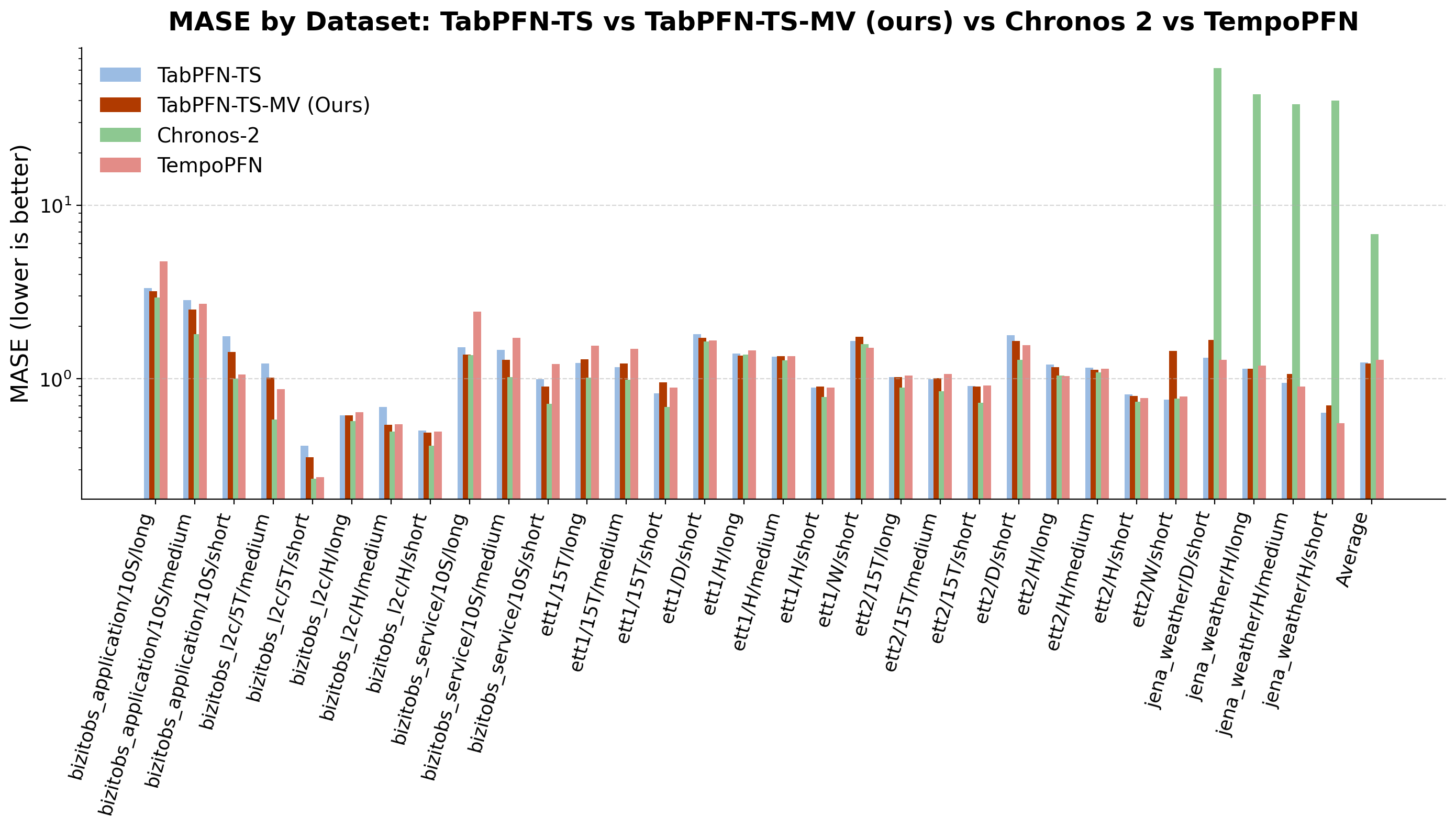}
    \caption{\textbf{Performance of TabPFN-TS-MV compared to other SOTA time series approaches:} The y-axis is log scale, Tabpfn-TS-MV (ours) and Chronos 2 supports channel dependent (CD) multivariate predictions while TabPFN-TS and TempoPFN only support univariate (channel independent) predictions.}
    \label{fig:mase_chronos}
\end{figure}

\begin{figure}[ht]
    \centering
    \includegraphics[width=\linewidth]{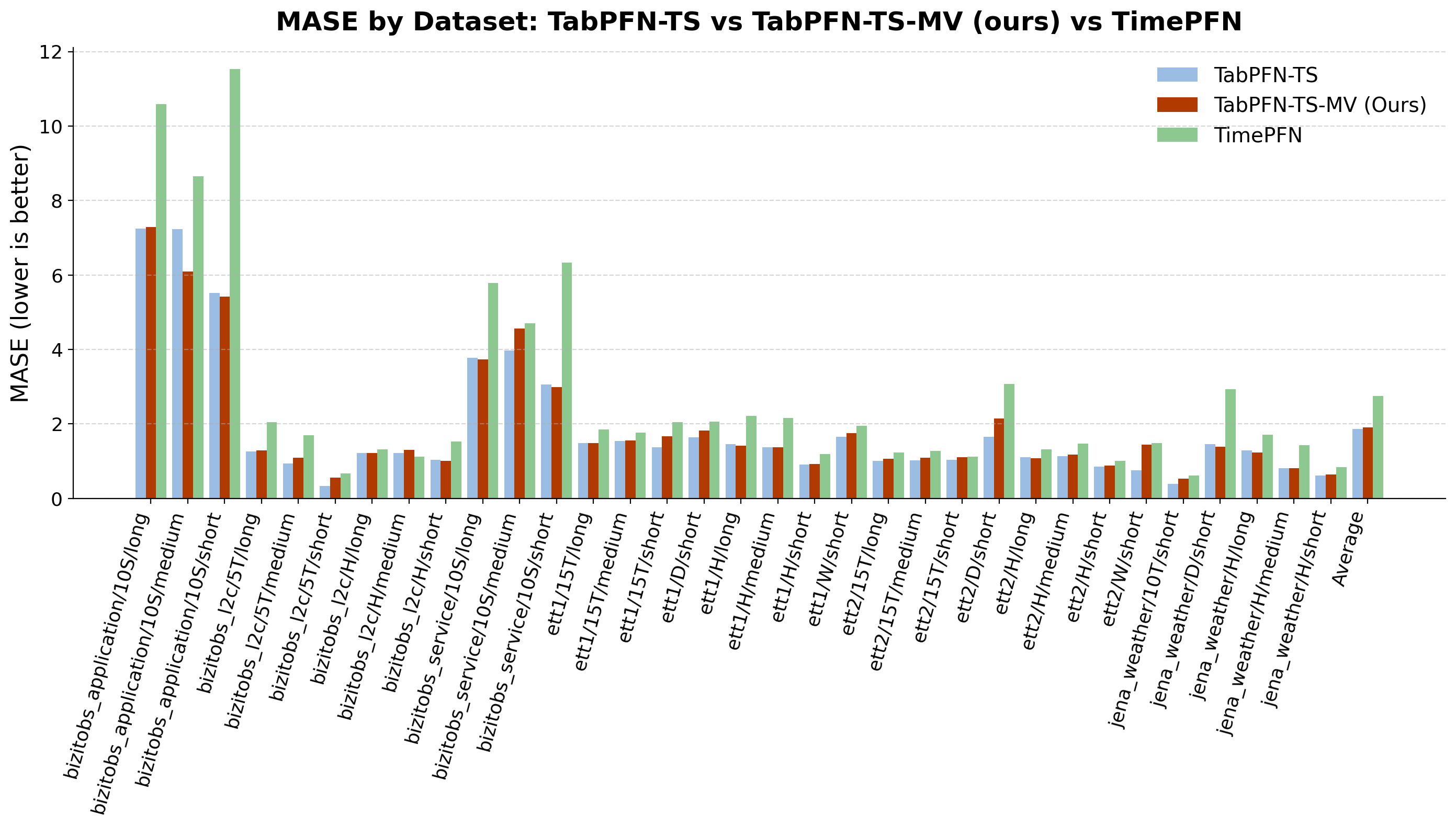}
    \caption{\textbf{Performance of TabPFN-TS-MV compared to TimePFN:} While TimePFN supports joint multivariate time series forecasting it only supports a maximum context window of 96 time steps. This poses a significant limitation as a majority of the datasets in Gift-Eval consist of much longer time series. All approaches are limited to a context window of 96 time steps (per variate) for fair comparison}
    \label{fig:mase_timepfn}
\end{figure}

A detailed per dataset MASE comparison is provided below (see Figure \ref{fig:mase_chronos}). A separate detailed comparison is provided for TimePFN, which requires a fixed context window of 96 time steps. We restrict our model's context to match this constraint for fairness (see Figure \ref{fig:mase_timepfn}).

\clearpage
\section{Investigating strategies for mitigating distributional shifts between channels}
\label{sec:mitigation}

\begin{figure}[ht]
    \centering
    \includegraphics[width=\linewidth]{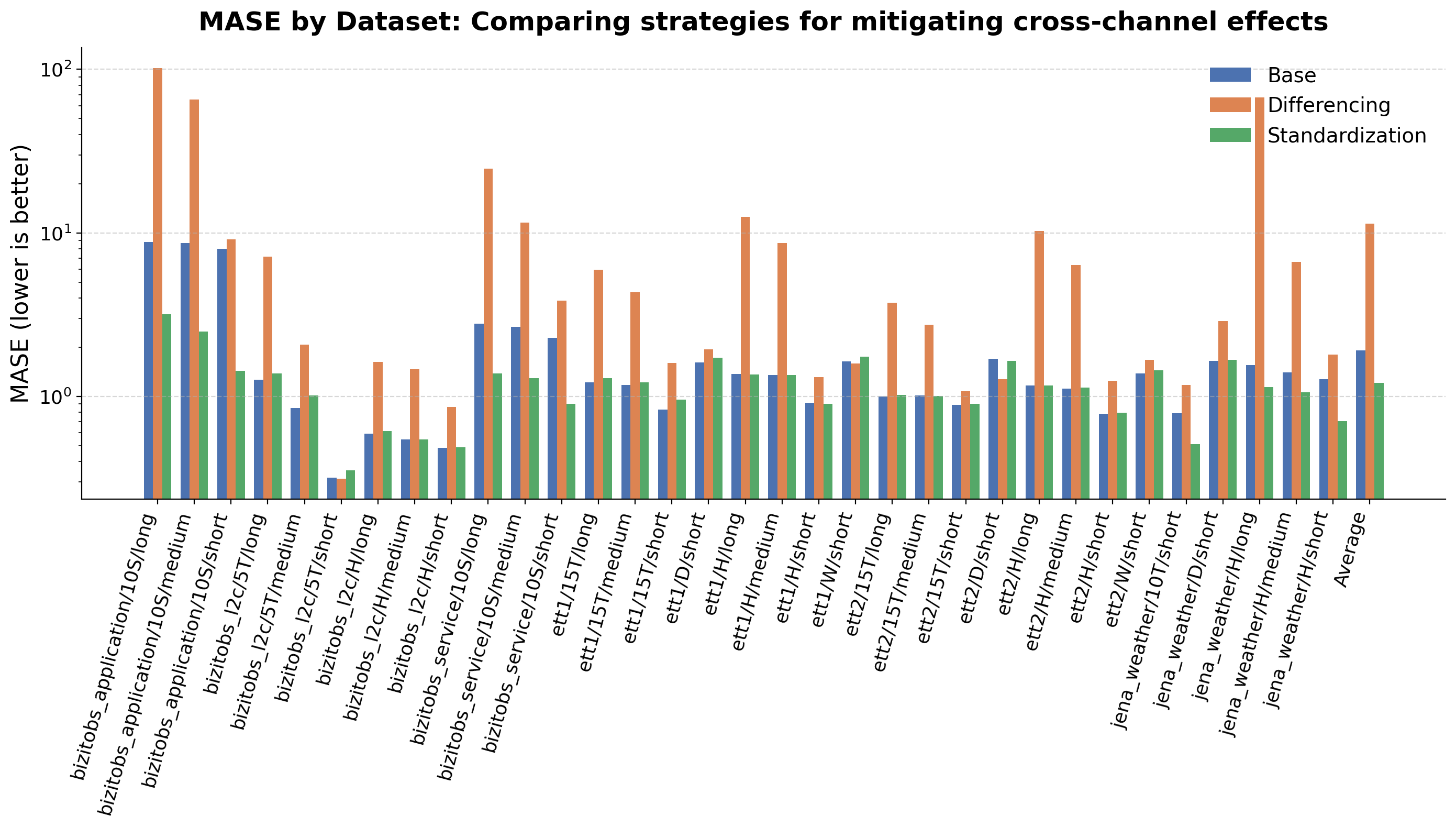}
    \caption{\textbf{Performance comparison between different strategies for mitigating cross-channel distributional shifts.} The y-axis is in log scale. We observe that while the standardizing approach greatly improves performance, first order differencing degrades performance. We incorporate the standardization approach into our framework.}
    \label{fig:mitigation}
\end{figure}

As observed in \cite{han2024capacity}, distributional differences between channels pose a significant challenge when performing joint multivariate forecasting. To address this, we evaluate two strategies for mitigating distributional discrepancies. First, we employ first-order differencing, where the model predicts the increment $\Delta y_t = y_t - y_{t-1}$ rather than the absolute value $y_t$. Second, we experiment with channel-wise standardization, training the model to predict normalized values $z_t = \frac{y_t - \mu}{\sigma}$ . Crucially, these operations are performed independently for each channel, and the inverse transformations are applied to the predictions prior to calculating evaluation metrics. Our experiments reveal that while standardization improves accuracy, differencing actually degrades performance. The relative performance of these approaches is illustrated in Figure \ref{fig:mitigation}. Consequently, we adopt standardization for our final framework.

\end{document}